\documentclass[10pt,twocolumn]{article}

\usepackage[T1]{fontenc}
\usepackage{amsmath}
\usepackage{amssymb}
\usepackage{newtxtext}
\usepackage{newtxmath}
\usepackage{helvet}
\usepackage{courier}
\usepackage[letterpaper,margin=0.8in,columnsep=0.3in]{geometry}
\usepackage[hyphens]{url}
\usepackage{graphicx}
\usepackage{booktabs}
\usepackage{multirow}
\usepackage{natbib}
\usepackage{caption}
\usepackage{enumitem}
\usepackage{placeins}
\usepackage{needspace}
\usepackage{titlesec}
\usepackage{titling}
\usepackage{flushend}
\usepackage[hidelinks]{hyperref}
\pretitle{\begin{center}\fontsize{16}{18}\selectfont\bfseries}
\posttitle{\par\end{center}\vspace{-0.35em}}
\preauthor{\begin{center}\normalsize}
\postauthor{\par\end{center}\vspace{-2.1em}}
\titleformat{\section}{\large\bfseries}{}{0pt}{}
\titleformat{\subsection}{\normalsize\bfseries}{}{0pt}{}
\titlespacing*{\section}{0pt}{1.4ex plus 0.3ex minus 0.2ex}{0.6ex}
\titlespacing*{\subsection}{0pt}{1.1ex plus 0.2ex minus 0.2ex}{0.4ex}
\setlist[itemize]{leftmargin=*,topsep=2pt,itemsep=1pt,parsep=0pt,partopsep=0pt}
\title{Robust Multi-View Classification under Noisy Supervision\\
via Global Anchor Consensus}
\author{
Yuliang Yang$^{1}$, Hongzhe Zhang$^{2}$, Huiru Wang$^{1,*}$\\[0.4em]
\small $^{1}$College of Science, Beijing Forestry University, Beijing 100083, China\\
\small $^{2}$School of Software, North University of China, Taiyuan, Shanxi 030051, China\\
\small \texttt{yangyuliang@bjfu.edu.cn}\quad
\texttt{23130408492@st.nuc.edu.cn}\quad
\texttt{whr2019@bjfu.edu.cn}\\
\small $^{*}$Corresponding author
}
\date{}

\hypersetup{
  pdftitle={Robust Multi-View Classification under Noisy Supervision via Global Anchor Consensus},
  pdfauthor={Yuliang Yang, Hongzhe Zhang, Huiru Wang}
}

\begin{document}

\maketitle

\begin{abstract}
In recent years, multi-view learning has attracted increasing attention, as it integrates the complementary information of heterogeneous views. Most existing multi-view classification methods rely on accurate annotations to guarantee performance. However, noisy labels are ubiquitous in practice due to imperfect annotation, and the refinement signals that existing methods derive from models trained on such noisy supervision can gradually lose their reliability. To deal with this problem, we propose a novel Global Anchor-based Label Auditing method (GALA) for multi-view classification to resist the negative impact of noisy labels. Specifically, we construct a global anchor for each class in every view, which aggregates the samples of the whole class and thus offers a stable reference insensitive to individual predictions. Then, each view measures how close an instance is to the anchor of its observed label relative to the nearest competing anchor, and the per-view evaluations are fused with the classifier confidence into a cross-view audit score. Based on the audit scores, suspicious samples are assigned small weights, and an adaptive correction strategy rewrites a label only when the anchor-based candidate agrees with the classifier prediction. Finally, the corrected labels in turn refine the anchors and supervise noise-robust representation learning. Extensive experiments on six datasets demonstrate that GALA outperforms eight state-of-the-art methods, especially under high noise rates.
\end{abstract}

% ============================================================
% Introduction
% ============================================================
\section{Introduction}

In recent years, multi-view data have become ubiquitous in real-world applications, where the same object is described by several heterogeneous sources or feature extractors \citep{li2019survey,yan2021review}. For example, an image can be described by texture, shape, and color descriptors, and a social-media post may contain textual, visual, and acoustic content. By integrating the consistent and complementary information among views, multi-view learning has achieved promising performance in many tasks, such as classification \citep{zhang2019cpm,zhang2022deep,liu2023safe}, clustering \citep{lin2021completer,jin2023deep,yan2023gcfagg}, and retrieval \citep{yan2020deep,hu2021learning,feng2023rono}.

To further improve the reliability of multi-view decisions, recent studies estimate view-wise uncertainty or resolve the conflicts among views \citep{han2021trusted,xu2024reliable,duan2025fuml,liu2025enhancing}, and handle incomplete data by recovering missing views or modeling their uncertainty \citep{lin2022dual,xie2023exploring}. These methods, however, implicitly assume that the training labels are accurate. In practice, manual annotation is expensive, subjective, and error-prone, so noisy labels are unavoidable.

More importantly, annotation errors are rarely random. Annotators tend to mislabel the instances that are genuinely ambiguous, so realistic label noise is instance-dependent noise (IDN) \citep{xia2020part,chen2021beyond}. Since an ambiguous instance is usually assigned to a class that it truly resembles, its wrong label looks plausible and is easy to fit. As training proceeds, deep networks gradually memorize such corrupted supervision and produce predictions that agree with the wrong labels \citep{zhang2017understanding,arpit2017closer}.

\begin{figure}[t]
\centering
\includegraphics[width=0.92\columnwidth]{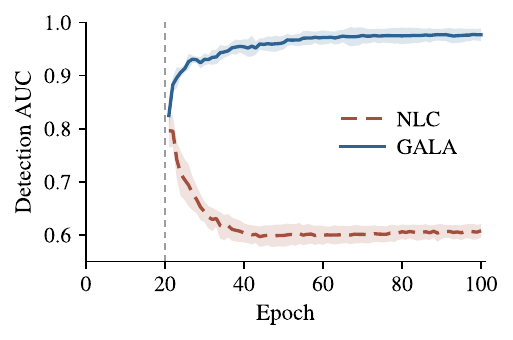}
\caption{Detection AUC of label-auditing evidence on Mfeat under 50\% IDN. The dashed line marks the end of the shared 20-epoch warm-up. Starting from comparable quality, NLC's neighbor-based evidence deteriorates, whereas GALA's anchor-based evidence improves.}
\label{fig:motivation}
\end{figure}

To alleviate the influence of noisy labels, several robust multi-view classification methods have been proposed. LNRMC \citep{liang2023label} designs a noise-robust regression model, TMNR \citep{xu2024trusted} learns view-specific noise correlation matrices that align the aggregated opinions with the noisy labels, and NLC \citep{xu2025noisy} detects and corrects noisy labels according to the predictions of nearest neighbors in the fused representation space. However, the signals that these methods rely on are produced by the model trained on the same observed labels. Taking NLC as an example, once the network begins to fit the corrupted labels, the neighboring predictions also drift toward the wrong labels and become less useful for detecting them. Consistent with this concern, Fig.~\ref{fig:motivation} shows that the neighbor-based detection AUC of NLC declines after the warm-up stage and never recovers. This motivates us to audit the observed labels with a reference that is more stable than the predictions of individual samples.

\begin{figure*}[t]
\centering
\includegraphics[width=\textwidth]{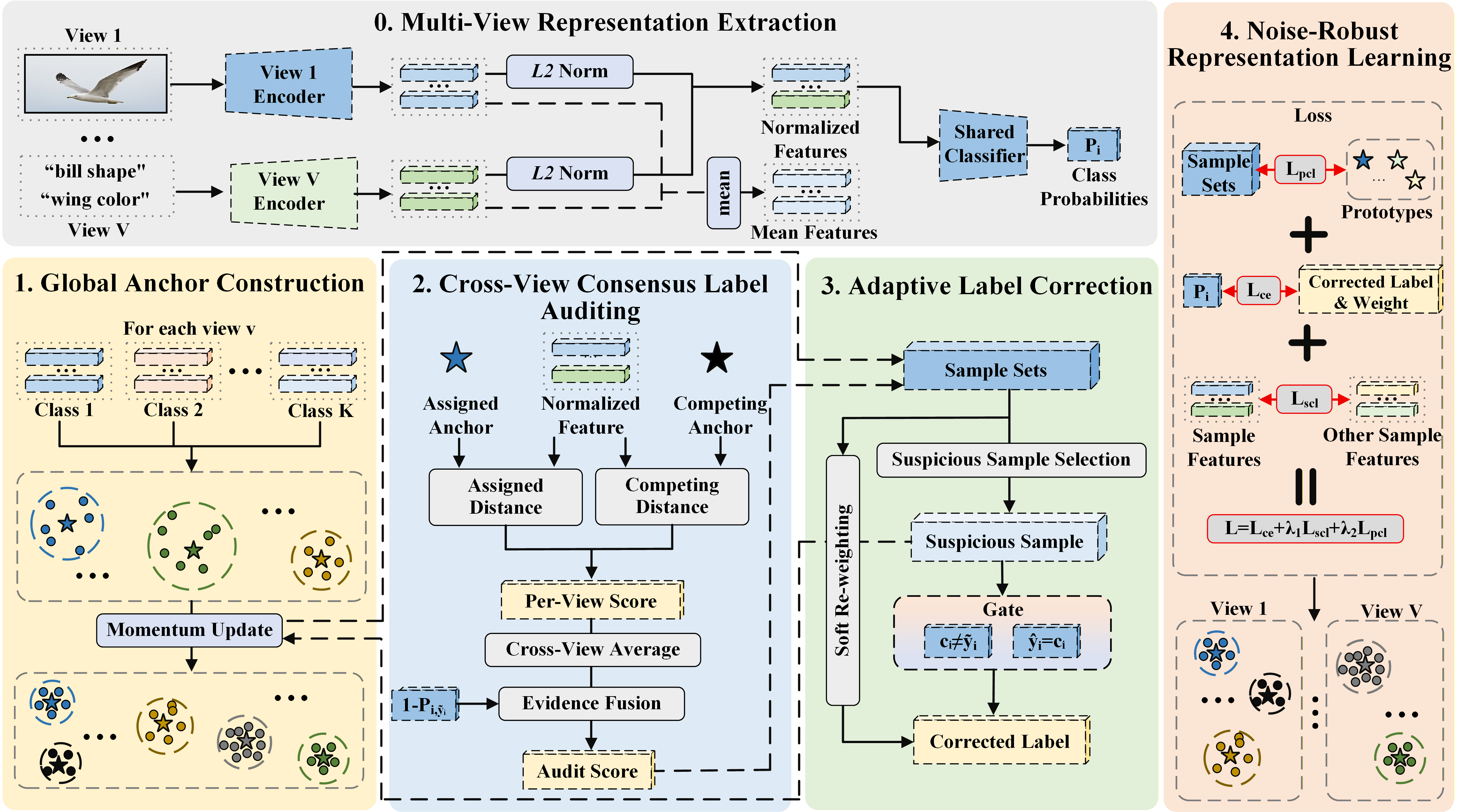}
\caption{The framework of the proposed GALA. GALA comprises global anchor construction, cross-view consensus label auditing, adaptive label correction, and noise-robust representation learning. View-specific class anchors are initialized by class-wise aggregation and updated with low-risk samples. GALA compares each instance with its assigned and competing anchors across views, fuses the resulting evidence with classifier confidence, and uses the audit score for soft re-weighting and conservative label correction. The corrected labels and weights supervise weighted classification, supervised contrastive learning, and prototypical contrastive learning.}
\label{fig:framework}
\end{figure*}

In this paper, we propose a Global Anchor-based Label Auditing method (GALA) for multi-view classification with noisy labels. Our key idea is that a global class anchor, which aggregates the representations of a whole class in each view, offers a reference that no individual prediction can easily distort. Specifically, as illustrated in Fig.~\ref{fig:framework}, GALA constructs a global anchor for each class in every view, and measures how close an instance is to the anchor of its observed label relative to the nearest competing anchor. The per-view evaluations are then averaged and fused with the classifier confidence to form a cross-view audit score. Based on the audit scores, suspicious samples are assigned small weights, and a label is rewritten only when the anchor-based candidate agrees with the classifier prediction. Finally, the corrected labels in turn refine the anchors and supervise noise-robust representation learning, so that the label quality and the representations improve together during training. The main contributions of this paper are summarized as follows:
\begin{itemize}
    \item To the best of our knowledge, we propose GALA as the first work to audit noisy labels in multi-view classification by leveraging view-specific global class anchors and cross-view consensus.
    \item We design an adaptive label correction strategy that couples soft sample weighting with a conservative rewriting rule, and alternately refines the noisy labels and the global anchors during training.
    \item Extensive experiments on six datasets show that GALA outperforms eight state-of-the-art methods, ranking first in 33 of 36 settings and surpassing the strongest baseline by 7.34 percentage points on average under 50\% noise.
\end{itemize}

% ============================================================
% Related Work
% ============================================================
\section{Related Work}

\subsection{Multi-View Learning}
Multi-view learning methods exploit the consistency and complementarity among views to obtain a more comprehensive description of the data \citep{zhang2019cpm,zhang2022deep,yan2023gcfagg}. One line of research focuses on improving the reliability of multi-view decisions. TMC \citep{han2021trusted} estimates view-specific uncertainty and dynamically integrates the resulting opinions, while ETMC \citep{han2022trusted} extends this idea through dynamic evidential fusion. To handle disagreement among views, ECML \citep{xu2024reliable} introduces a conflict-aware opinion aggregation strategy, FUML \citep{duan2025fuml} models view-specific reliability through fuzzy memberships, and \citet{liu2025enhancing} develop an adaptive rejection mechanism for unreliable decisions. Another line of research addresses incomplete observations. DCP \citep{lin2022dual} recovers missing views through dual contrastive prediction under an information-theoretic framework, whereas UIMC \citep{xie2023exploring} models the uncertainty introduced during view imputation. More recently, NCPD-TSMVC combines neural-collapse-inspired prototype priors with evidential fusion to improve representation learning and prediction reliability under limited supervision \citep{guo2026neural}. Although these methods improve robustness to view uncertainty, cross-view conflict, missing observations, or limited labels, none addresses the central problem considered here: corrupted supervision in multi-view classification.

\subsection{Learning with Noisy Labels}
Label noise is commonly divided into class-conditional noise (CCN) and instance-dependent noise (IDN) \citep{song2023learning}. Under CCN, the corruption is independent of the instance features given the true class and can therefore be addressed by estimating a class-transition matrix for loss correction \citep{patrini2017making}. Under IDN, by contrast, the probability of mislabeling depends on the instance itself, making ambiguous samples more likely to be corrupted; this setting more closely reflects annotation errors in practice \citep{xia2020part,chen2021beyond}. We therefore focus on IDN. Existing approaches include robust sample selection, semi-supervised learning, and input interpolation. Co-teaching \citep{han2018coteaching} trains two peer networks that exchange small-loss samples, DivideMix \citep{li2020dividemix} separates clean and noisy examples through loss modeling and learns from the latter in a semi-supervised manner, and MixUp \citep{zhang2018mixup} reduces overfitting to corrupted labels by interpolating both inputs and targets.

Class-level prototypes provide another way to reduce reliance on individual noisy examples. CleanNet \citep{lee2018cleannet} verifies labels against class reference embeddings but requires manually verified examples for a subset of classes. Deep Self-Learning \citep{han2019deep} extracts multiple prototypes for each class and uses them to refine noisy supervision, while MoPro \citep{li2021mopro} maintains momentum prototypes for online label correction in webly supervised learning. More recent studies have extended prototype-based robustness to pretrained vision-language models and graph learning. NLPrompt treats text embeddings as class prototypes and uses prototype-guided optimal transport to separate clean from noisy examples for robust prompt learning \citep{pan2025nlprompt}. Prototype-Guided Supervision instead constructs class-level semantic anchors from node representations and replaces direct supervision from potentially corrupted node labels with prototype-level supervision \citep{li2026prototype}. These studies further demonstrate the value of class-level structure under noisy supervision, but they are tailored to vision-language adaptation or graph learning. They neither maintain a separate class-anchor space for each view nor exploit agreement across views to audit observed labels, leaving the problem addressed by GALA unresolved.

\subsection{Multi-View Learning with Noisy Labels}
Noisy supervision has been studied in cross-modal retrieval \citep{hu2021learning,feng2023rono} and in incomplete multi-view multi-label learning \citep{wang2025dynamic}, but only a few methods target multi-view classification.

LNRMC \citep{liang2023label} builds a label-noise robust regression model with robust losses, without explicitly separating clean and corrupted samples. TMNR \citep{xu2024trusted} constructs evidential opinions for each view and learns view-specific noise correlation matrices that align the aggregated opinion with the noisy labels; the matrices are themselves fitted to the corrupted supervision and provide no per-sample auditing. NLC \citep{xu2025noisy} detects a noisy label by comparing it with the predictions of its nearest neighbors in the fused representation space and relabels flagged samples by neighbor voting; since the neighboring predictions are produced by the model trained on the same observed labels, this evidence can weaken once the network fits the corrupted labels.

GALA differs from single-view prototype methods in that it maintains an anchor set for each view and audits labels through cross-view agreement. It also differs from NLC: the auditing evidence comes from class-level anchors rather than the predictions of individual neighbors, and a conservative rule changes a label only when the anchor candidate agrees with the classifier.

% ============================================================
% Method
% ============================================================
\section{Method}

\subsection{Problem Formulation}

Consider a multi-view training set $\mathcal{D}=\{(\mathbf{x}_i,\tilde{y}_i)\}_{i=1}^{N}$ of $N$ instances, where each instance $\mathbf{x}_i=\{\mathbf{x}_i^v\}_{v=1}^{V}$ is observed in $V$ views and $\tilde{y}_i\in\{1,\dots,K\}$ denotes its observed label over $K$ classes. Since manual annotation is error-prone, $\tilde{y}_i$ may disagree with the latent ground-truth label. We aim to train a multi-view classifier that remains accurate on test data despite such corrupted supervision.

For view $v$, an encoder $f^v(\cdot)$ maps $\mathbf{x}_i^v$ to a representation $\mathbf{z}_i^v=f^v(\mathbf{x}_i^v)\in\mathbb{R}^d$, where $d$ is the dimension of the latent space. We denote its $\ell_2$-normalized form by $\bar{\mathbf{z}}_i^v$. The fused representation is the average of all view-specific representations, i.e., $\mathbf{z}_i=\frac{1}{V}\sum_{v=1}^{V}\mathbf{z}_i^v$. A shared classifier $g(\cdot)$ outputs the class-probability vector $\mathbf{p}_i=\mathrm{softmax}(g(\mathbf{z}_i))$, where $p_{i,c}$, the $c$-th element of $\mathbf{p}_i$, denotes the predicted probability that instance $i$ belongs to class $c$.

\subsection{Global Anchor Construction}

GALA summarizes class-level structure with global anchors rather than consulting individual neighbors. Exploiting the early-learning behavior of DNNs \citep{arpit2017closer}, we warm up the model with standard cross-entropy before initializing the anchor of class $c$ in view $v$ as the normalized class mean:
\begin{equation}
\mathbf{a}_c^v=\frac{\sum_{i\in\mathcal{N}_c}\bar{\mathbf{z}}_i^v}{\big\|\sum_{i\in\mathcal{N}_c}\bar{\mathbf{z}}_i^v\big\|_2},\quad
\mathcal{N}_c=\{i\mid \tilde{y}_i=c\}.
\label{eq:anchor_init}
\end{equation}
Initialization uses the observed labels, but each anchor pools representations across a class and is therefore less sensitive to any individual sample. The anchor remains useful provided that corrupted samples do not dominate the class.

After warm-up, each anchor is updated by momentum using the samples judged as clean:
\begin{equation}
\mathbf{a}_c^v \leftarrow \mu\,\mathbf{a}_c^v+(1-\mu)\,\frac{1}{|\mathcal{C}_c|}\sum_{i\in\mathcal{C}_c}\bar{\mathbf{z}}_i^v,
\label{eq:anchor_update}
\end{equation}
followed by $\ell_2$ normalization, where $\mathcal{C}_c$ contains the clean samples with corrected label $c$ in the current mini-batch and $\mu$ is the momentum coefficient. If $\mathcal{C}_c$ is empty, the corresponding anchor remains unchanged. The warm-up stage and the conservative correction rule introduced below stabilize the feedback between label correction and anchor refinement.

\subsection{Cross-View Consensus Label Auditing}

An observed label is suspicious if the corresponding instance lies far from the anchor of its labeled class (the assigned anchor) but close to a competing anchor, consistently across views. Since both representations and anchors are normalized, we measure their distance by
\begin{equation}
d_{i,c}^v=1-\langle\bar{\mathbf{z}}_i^v,\mathbf{a}_{c}^v\rangle,
\label{eq:anchor_distance}
\end{equation}
where $\langle\cdot,\cdot\rangle$ denotes the inner product. To account for relative class separation, each view contrasts the assigned anchor with the nearest competing one:
\begin{equation}
s_i^v=
\frac{d_{i,\tilde{y}_i}^v}
{d_{i,\tilde{y}_i}^v+\min_{c\neq\tilde{y}_i}d_{i,c}^v+\delta},
\label{eq:per_view_score}
\end{equation}
where $\delta=10^{-8}$ is a small constant for numerical stability. This ratio places the view-specific scores on a common bounded scale. We then average the scores across views and combine them with the classifier confidence:
\begin{equation}
s_i=\frac{\alpha}{V}\sum_{v=1}^{V}s_i^v+(1-\alpha)\big(1-p_{i,\tilde{y}_i}\big),
\label{eq:audit_score}
\end{equation}
where $\alpha$ balances the anchor-based and classifier-based evidence. After warm-up, corrected labels and soft weights refine the classifier, whose confidence complements the anchor score. A larger $s_i$ indicates a higher risk that $\tilde{y}_i$ is corrupted.

\subsection{Adaptive Label Correction}

The audit score controls both sample re-weighting and label correction. Each sample receives a soft weight
\begin{equation}
w_i=\max(1-s_i,\ \epsilon),
\label{eq:weight}
\end{equation}
where $\epsilon$ avoids zero weights. Down-weighting suppresses unreliable supervision but cannot recover the potentially useful class information of a corrupted sample. We therefore derive an anchor-based candidate by aggregating similarities across views:
\begin{equation}
\hat{y}_i=\arg\max_{c}\sum_{v=1}^{V}\langle\bar{\mathbf{z}}_i^v,\mathbf{a}_c^v\rangle.
\label{eq:pseudo}
\end{equation}
Within a mini-batch of size $B$, the $\lceil\eta B\rceil$ largest scores are flagged as suspicious, where $\eta$ controls the correction budget. We denote the smallest flagged score by $\theta_\eta$ and the classifier prediction by $c_i=\arg\max_c p_{i,c}$. The corrected label is
\begin{equation}
\bar{y}_i=
\begin{cases}
\hat{y}_i, & \text{if}\ s_i\geq\theta_\eta,\ c_i\neq\tilde{y}_i,\ \text{and}\ \hat{y}_i=c_i,\\[2pt]
\tilde{y}_i, & \text{otherwise}.
\end{cases}
\label{eq:correction}
\end{equation}
When $\eta=0$, no sample is flagged as suspicious, and all observed labels remain unchanged. A label is rewritten only when the sample is suspicious and the classifier agrees with the anchor-based candidate on an alternative class. This conservative rule avoids injecting an additional error, while an uncorrected noisy label is still down-weighted by Eq.~(\ref{eq:weight}). The lowest-scoring $(1-\eta)$ fraction is treated as clean for the anchor update in Eq.~(\ref{eq:anchor_update}).

\subsection{Noise-Robust Representation Learning}

The corrected labels and soft weights supervise both classification and representation learning. We train the classifier with weighted cross-entropy:
\begin{equation}
\mathcal{L}_{\mathrm{ce}}=\frac{\sum_{i}w_i\,\ell_{\mathrm{ce}}(\mathbf{p}_i,\bar{y}_i)}{\sum_{i}w_i},
\label{eq:wce}
\end{equation}
where $\ell_{\mathrm{ce}}$ is the cross-entropy loss. We also use a weighted supervised contrastive loss \citep{khosla2020supervised}. Let $\ell_{\mathrm{sup}}(\bar{\mathbf{z}}_i^v,\bar{y}_i)$ denote the standard supervised contrastive loss over all view-specific features in the mini-batch, where features sharing the same corrected label form positive pairs. We define
\begin{equation}
\mathcal{L}_{\mathrm{scl}}
=\frac{1}{V\sum_i w_i}
\sum_{i=1}^{B}\sum_{v=1}^{V}
w_i\,\ell_{\mathrm{sup}}(\bar{\mathbf{z}}_i^v,\bar{y}_i).
\label{eq:scl}
\end{equation}

A prototypical contrastive loss keeps features compact around their anchors:
\begin{equation}
\mathcal{L}_{\mathrm{pcl}}=\frac{1}{V}\sum_{v=1}^{V}\frac{1}{B}\sum_{i=1}^{B}\Big[w_i\,\ell_i^v(\bar{y}_i)+(1-w_i)\,\ell_i^v(\hat{y}_i)\Big],
\label{eq:pcl}
\end{equation}
where $\ell_i^v(y)$ is the cross-entropy loss for target $y$ computed from the anchor-similarity logits $\langle\bar{\mathbf{z}}_i^v,\mathbf{a}_c^v\rangle/\tau$ over classes $c$, and $\tau$ is the temperature. Thus, reliable samples follow their corrected labels, whereas suspicious samples rely more on the anchor-based candidates.

% ============================================================
% Experiments
% ============================================================
% Declare the main table before the section so that a two-column float can
% appear at the top of the page where the experiments begin.

\subsection{Overall Loss}

The overall training objective of GALA is
\begin{equation}
\mathcal{L}=\mathcal{L}_{\mathrm{ce}}+\lambda_1\,\mathcal{L}_{\mathrm{scl}}+\lambda_2\,\mathcal{L}_{\mathrm{pcl}},
\label{eq:overall}
\end{equation}
where $\lambda_1$ and $\lambda_2$ are balance parameters. During warm-up, GALA uses only standard cross-entropy on the observed labels. Label auditing, correction, anchor updating, and the full objective in Eq.~(\ref{eq:overall}) are activated afterwards. The anchors are updated without gradient computation.

\section{Experiments}

\subsection{Experimental Setup}

\paragraph{Datasets.}
We evaluate GALA on six multi-view datasets: LandUse21, Leaves100, Scene15, BBC, Mfeat, and Digit4k. \textbf{LandUse21} contains 2,100 samples from 21 classes with three views; \textbf{Leaves100} contains 1,600 samples from 100 classes with three views; \textbf{Scene15} contains 4,485 samples from 15 classes with three views; \textbf{BBC} contains 685 documents from five classes with four views; \textbf{Mfeat} contains 2,000 handwritten-digit samples from 10 classes with six views; and \textbf{Digit4k} contains 4,000 samples from four classes with two views. Following the IDN protocol of TMNR \citep{xu2024trusted}, we randomly split each dataset into 80\% training and 20\% testing samples and corrupt the training labels at rates of $\{0,0.1,0.2,0.3,0.4,0.5\}$, where 0 corresponds to training with clean labels. The protocol assigns feature-dependent corruption probabilities and maps a selected sample to its most confusable class. The test labels are always kept clean.

\paragraph{Compared Methods.}
We compare GALA with eight representative methods. DCCAE \citep{wang2015deep} learns correlated representations with deep canonically correlated autoencoders, and DCP \citep{lin2022dual} enforces cross-view consistency through dual contrastive prediction. TMC \citep{han2021trusted}, ETMC \citep{han2022trusted}, ECML \citep{xu2024reliable}, and UIMC \citep{xie2023exploring} improve the reliability of multi-view decisions through uncertainty estimation, dynamic evidential fusion, conflictive opinion aggregation, and uncertainty-aware imputation, respectively. TMNR \citep{xu2024trusted} and NLC \citep{xu2025noisy} are designed for multi-view learning with noisy labels: TMNR models the corruption process with view-specific noise correlation matrices over evidential opinions, while NLC verifies and calibrates labels with the predictions of reliable neighbors. For a fair comparison, all compared methods use their official implementations with the parameters recommended in the original papers; if no recommendation is available, we carefully tune them. All methods are evaluated with identical data splits and random seeds.

\paragraph{Implementation Details.}
GALA is implemented in PyTorch 2.5.1 and trained on an NVIDIA GeForce RTX 4090. Each view has an independent MLP encoder with batch normalization, ReLU, and 0.3 dropout, producing a 128-dimensional representation; mean fusion and a shared linear classifier follow. We train for 100 epochs with a batch size of 128, using the Adam optimizer with an initial learning rate of $10^{-3}$, a weight decay of $10^{-4}$, and a cosine annealing schedule; the first 20 epochs serve as warm-up. For all datasets, $\alpha=0.5$, $\mu=0.9$, $\tau=0.1$, $\epsilon=0.05$, and $\lambda_1=\lambda_2=0.5$. The correction budget $\eta$ is set to the known noise rate, which can otherwise be estimated from noisy training data \citep{garg2024noise}. We report mean accuracy and standard deviation over five random seeds.

\begin{table*}[!t]
\centering
\small
\tabcolsep=1pt
\begin{tabular*}{\textwidth}{@{\extracolsep{\fill}}c|c|cccccccc|c}
\hline
Dataset & Noise & DCCAE & TMC & ETMC & DCP & UIMC & ECML & TMNR & NLC & GALA \\
Year & & ICML'15 & ICLR'21 & TPAMI'23 & TPAMI'23 & CVPR'23 & AAAI'24 & IJCAI'24 & AAAI'25 & Ours \\
\hline
\multirow{6}{*}{LandUse21}
& 0\%  & 32.62$\pm$1.74 & 49.62$\pm$3.26 & 45.00$\pm$3.04 & 71.38$\pm$2.29 & 68.00$\pm$1.58 & 37.95$\pm$1.94 & 41.71$\pm$2.46 & \textbf{75.19$\pm$1.19} & \underline{73.29$\pm$0.96} \\
& 10\% & 30.48$\pm$1.23 & 48.71$\pm$4.10 & 43.33$\pm$3.92 & 55.10$\pm$0.82 & 54.10$\pm$1.10 & 34.10$\pm$2.43 & 39.00$\pm$2.91 & \textbf{70.14$\pm$1.92} & \underline{69.90$\pm$1.50} \\
& 20\% & 29.33$\pm$1.53 & 46.48$\pm$2.49 & 41.33$\pm$2.75 & 44.67$\pm$2.39 & 47.24$\pm$1.27 & 31.52$\pm$1.92 & 35.95$\pm$2.27 & \underline{65.52$\pm$1.75} & \textbf{68.43$\pm$2.10} \\
& 30\% & 29.24$\pm$1.92 & 43.24$\pm$3.54 & 39.76$\pm$2.76 & 34.00$\pm$1.94 & 40.14$\pm$1.45 & 30.43$\pm$1.39 & 36.71$\pm$2.28 & \underline{57.76$\pm$3.00} & \textbf{65.29$\pm$0.92} \\
& 40\% & 26.00$\pm$1.59 & 40.10$\pm$1.76 & 37.19$\pm$2.89 & 27.90$\pm$1.14 & 28.62$\pm$1.03 & 26.19$\pm$1.68 & 33.10$\pm$1.70 & \underline{52.24$\pm$2.48} & \textbf{62.76$\pm$2.43} \\
& 50\% & 26.29$\pm$0.98 & 36.57$\pm$2.34 & 35.24$\pm$2.45 & 21.86$\pm$1.75 & 25.48$\pm$0.77 & 25.43$\pm$1.60 & 29.38$\pm$3.07 & \underline{48.29$\pm$2.41} & \textbf{59.48$\pm$2.42} \\
\hline
\multirow{6}{*}{Leaves100}
& 0\%  & 62.00$\pm$4.17 & 94.19$\pm$0.32 & 90.56$\pm$1.29 & 97.25$\pm$0.67 & 97.06$\pm$0.47 & 73.19$\pm$1.46 & 73.75$\pm$2.55 & \underline{98.50$\pm$0.46} & \textbf{99.88$\pm$0.25} \\
& 10\% & 61.31$\pm$2.50 & 93.25$\pm$0.61 & 89.25$\pm$1.20 & 84.56$\pm$1.50 & 92.06$\pm$0.51 & 73.00$\pm$2.15 & 68.88$\pm$2.17 & \underline{96.69$\pm$0.61} & \textbf{99.50$\pm$0.54} \\
& 20\% & 56.88$\pm$3.55 & 89.56$\pm$1.23 & 86.00$\pm$2.01 & 72.75$\pm$2.53 & 87.00$\pm$1.49 & 72.06$\pm$1.08 & 64.19$\pm$2.03 & \underline{93.25$\pm$1.93} & \textbf{99.50$\pm$0.38} \\
& 30\% & 55.19$\pm$2.69 & 85.38$\pm$1.65 & 81.56$\pm$2.30 & 59.25$\pm$3.54 & 80.44$\pm$1.56 & 67.88$\pm$1.50 & 60.94$\pm$2.04 & \underline{87.25$\pm$1.93} & \textbf{96.69$\pm$0.88} \\
& 40\% & 53.00$\pm$2.58 & 81.56$\pm$2.41 & 76.44$\pm$2.91 & 48.00$\pm$1.35 & 71.06$\pm$2.16 & 65.38$\pm$2.52 & 57.75$\pm$2.12 & \underline{82.25$\pm$2.64} & \textbf{94.75$\pm$1.51} \\
& 50\% & 50.19$\pm$2.18 & 73.81$\pm$3.61 & 68.50$\pm$2.47 & 39.75$\pm$2.14 & 65.50$\pm$1.02 & 59.25$\pm$1.46 & 55.63$\pm$2.91 & \underline{75.19$\pm$3.23} & \textbf{92.25$\pm$2.28} \\
\hline
\multirow{6}{*}{Scene15}
& 0\%  & 49.50$\pm$1.54 & 72.11$\pm$1.14 & 70.30$\pm$1.31 & 74.52$\pm$2.41 & 75.25$\pm$0.33 & 59.49$\pm$1.39 & 63.84$\pm$1.40 & \textbf{80.13$\pm$1.20} & \underline{79.82$\pm$0.88} \\
& 10\% & 48.23$\pm$1.76 & 72.02$\pm$1.28 & 69.99$\pm$1.25 & 53.07$\pm$2.24 & 74.18$\pm$1.51 & 58.66$\pm$0.97 & 63.34$\pm$1.20 & \underline{77.99$\pm$1.00} & \textbf{78.89$\pm$0.66} \\
& 20\% & 48.32$\pm$1.68 & 72.06$\pm$1.13 & 69.65$\pm$0.81 & 42.63$\pm$1.10 & 71.42$\pm$1.12 & 60.13$\pm$0.74 & 62.74$\pm$1.85 & \underline{76.54$\pm$1.00} & \textbf{77.86$\pm$0.21} \\
& 30\% & 47.42$\pm$1.34 & 71.82$\pm$1.09 & 69.21$\pm$1.48 & 33.89$\pm$1.27 & 69.94$\pm$1.44 & 59.26$\pm$1.30 & 59.40$\pm$1.96 & \underline{73.76$\pm$1.10} & \textbf{77.68$\pm$0.50} \\
& 40\% & 47.76$\pm$1.32 & 68.83$\pm$0.38 & 66.47$\pm$1.29 & 25.71$\pm$0.53 & 67.89$\pm$0.82 & 57.64$\pm$1.16 & 53.13$\pm$2.07 & \underline{70.77$\pm$1.87} & \textbf{77.32$\pm$0.86} \\
& 50\% & 46.58$\pm$2.23 & 65.15$\pm$1.72 & 63.63$\pm$2.27 & 19.60$\pm$1.60 & 63.61$\pm$2.34 & 54.49$\pm$0.94 & 52.13$\pm$3.53 & \underline{68.61$\pm$1.74} & \textbf{75.23$\pm$1.15} \\
\hline
\multirow{6}{*}{BBC}
& 0\%  & 86.42$\pm$1.94 & 95.77$\pm$1.42 & 95.77$\pm$1.49 & 91.68$\pm$2.10 & \underline{96.20$\pm$1.42} & 94.45$\pm$2.29 & 94.01$\pm$2.46 & 91.53$\pm$2.40 & \textbf{98.10$\pm$1.09} \\
& 10\% & 83.06$\pm$1.81 & 92.55$\pm$2.82 & 92.99$\pm$2.87 & 87.88$\pm$3.08 & \underline{93.87$\pm$1.76} & 90.22$\pm$2.64 & 91.09$\pm$3.25 & 88.17$\pm$4.04 & \textbf{96.20$\pm$1.49} \\
& 20\% & 82.19$\pm$1.88 & 92.55$\pm$2.03 & 92.55$\pm$1.56 & 83.21$\pm$3.13 & \underline{93.28$\pm$0.85} & 89.63$\pm$2.33 & 91.09$\pm$1.87 & 84.23$\pm$5.28 & \textbf{94.01$\pm$0.85} \\
& 30\% & 72.41$\pm$2.23 & 81.31$\pm$1.76 & \underline{82.33$\pm$2.46} & 74.45$\pm$4.71 & 80.58$\pm$2.79 & 79.12$\pm$2.83 & 80.44$\pm$2.59 & 75.47$\pm$4.86 & \textbf{87.01$\pm$1.56} \\
& 40\% & 68.61$\pm$2.01 & 75.33$\pm$1.98 & \underline{76.20$\pm$2.94} & 68.90$\pm$4.96 & 74.89$\pm$2.04 & 72.41$\pm$1.81 & 74.16$\pm$1.50 & 71.97$\pm$9.57 & \textbf{81.61$\pm$2.59} \\
& 50\% & 58.10$\pm$1.99 & 64.81$\pm$2.82 & \underline{64.96$\pm$2.53} & 60.00$\pm$3.04 & 63.21$\pm$1.35 & 63.06$\pm$1.35 & 63.06$\pm$2.79 & 56.05$\pm$3.84 & \textbf{70.51$\pm$4.44} \\
\hline
\multirow{6}{*}{Mfeat}
& 0\%  & 97.40$\pm$0.58 & 97.95$\pm$0.62 & 98.25$\pm$0.65 & \underline{98.80$\pm$0.19} & \underline{98.80$\pm$0.10} & 97.15$\pm$0.77 & 96.75$\pm$0.32 & 97.55$\pm$0.46 & \textbf{99.40$\pm$0.30} \\
& 10\% & 97.50$\pm$0.45 & 97.65$\pm$0.37 & 97.70$\pm$0.24 & 96.50$\pm$1.17 & \underline{97.95$\pm$0.33} & 96.35$\pm$0.37 & 96.55$\pm$0.51 & 97.20$\pm$0.51 & \textbf{98.70$\pm$0.24} \\
& 20\% & 96.70$\pm$0.33 & \underline{97.65$\pm$0.86} & 97.55$\pm$0.89 & 92.05$\pm$0.66 & \underline{97.65$\pm$0.60} & 96.35$\pm$0.97 & 96.50$\pm$0.57 & 96.85$\pm$0.60 & \textbf{98.25$\pm$0.35} \\
& 30\% & 95.60$\pm$0.85 & \underline{95.85$\pm$0.44} & 95.70$\pm$0.70 & 82.30$\pm$3.90 & \underline{95.85$\pm$0.56} & 94.30$\pm$0.89 & 95.25$\pm$0.71 & 95.55$\pm$1.08 & \textbf{97.85$\pm$0.77} \\
& 40\% & 91.95$\pm$1.07 & 94.30$\pm$0.51 & 94.25$\pm$0.16 & 76.45$\pm$0.62 & 93.95$\pm$0.76 & 92.85$\pm$0.68 & 93.60$\pm$0.51 & \underline{95.00$\pm$1.77} & \textbf{96.70$\pm$0.24} \\
& 50\% & 82.20$\pm$3.18 & 89.50$\pm$2.26 & \underline{90.05$\pm$1.34} & 62.05$\pm$1.58 & 88.00$\pm$2.16 & 87.80$\pm$2.68 & 86.80$\pm$2.35 & 89.55$\pm$3.43 & \textbf{92.10$\pm$1.69} \\
\hline
\multirow{6}{*}{Digit4k}
& 0\%  & \underline{91.10$\pm$0.39} & 88.70$\pm$0.76 & 88.98$\pm$0.64 & 88.85$\pm$2.09 & 90.35$\pm$0.46 & 88.62$\pm$0.69 & 87.52$\pm$0.78 & 90.20$\pm$1.06 & \textbf{91.75$\pm$0.50} \\
& 10\% & \underline{91.02$\pm$0.37} & 88.60$\pm$0.82 & 88.80$\pm$0.84 & 84.90$\pm$0.99 & 90.08$\pm$0.88 & 88.55$\pm$0.74 & 87.18$\pm$0.92 & 87.95$\pm$2.15 & \textbf{91.45$\pm$0.39} \\
& 20\% & \underline{90.83$\pm$0.33} & 89.27$\pm$0.72 & 89.40$\pm$0.75 & 76.43$\pm$2.98 & 89.55$\pm$0.74 & 89.05$\pm$0.70 & 88.00$\pm$0.84 & 86.65$\pm$2.82 & \textbf{91.10$\pm$0.63} \\
& 30\% & \underline{89.77$\pm$0.47} & 88.20$\pm$0.70 & 88.38$\pm$0.77 & 71.10$\pm$4.71 & 89.33$\pm$0.56 & 88.23$\pm$0.71 & 87.20$\pm$0.59 & 85.80$\pm$2.98 & \textbf{90.53$\pm$0.53} \\
& 40\% & 87.22$\pm$0.51 & \underline{87.72$\pm$0.59} & 87.65$\pm$0.39 & 68.20$\pm$9.41 & 87.47$\pm$0.41 & 87.60$\pm$0.42 & 86.87$\pm$0.45 & 84.75$\pm$2.76 & \textbf{89.10$\pm$0.37} \\
& 50\% & 79.17$\pm$1.05 & \underline{84.60$\pm$0.74} & 84.52$\pm$0.57 & 56.23$\pm$15.73 & 83.95$\pm$0.86 & 84.58$\pm$0.74 & 83.52$\pm$0.76 & 77.05$\pm$5.08 & \textbf{86.17$\pm$2.13} \\
\hline
\end{tabular*}
\normalsize
\caption{Classification accuracy (\%) of GALA and eight compared methods on six datasets with different proportions of instance-dependent noise. The ``Noise'' column denotes the percentage of corrupted training labels, with 0\% indicating clean training data. The best and second-best results are indicated by boldface and underlining, respectively.}
\label{tab:main_results}
\end{table*}

\subsection{Overall Performance}

Table~\ref{tab:main_results} reports the classification accuracy under different noise rates. GALA ranks first in 33 of the 36 settings and second in the remaining three, and its advantage grows as the noise rate increases. At 50\% noise, GALA outperforms the strongest competing method by 7.34 percentage points on average, and its average margin over NLC increases from 1.52 points on clean data to 10.17 points. On the clean datasets, GALA ranks first on four and second on the other two, showing that its robustness to label noise does not come at the cost of clean-data accuracy.

\subsection{Label Auditing Analysis}

\begin{figure*}[!t]
\centering
\includegraphics[width=\textwidth]{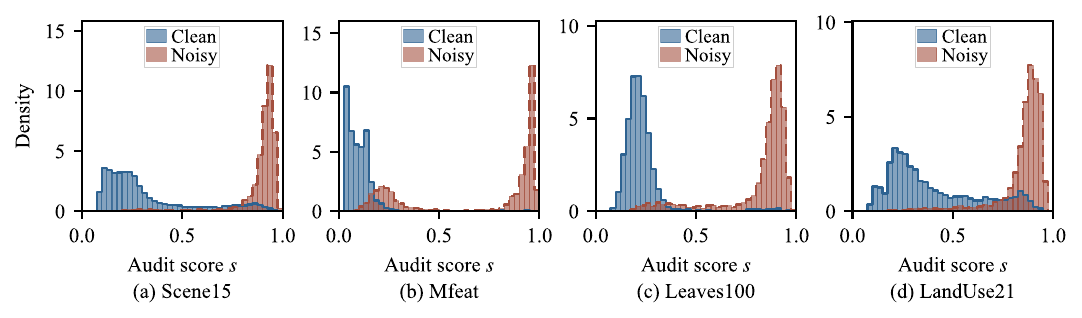}
\caption{Distributions of audit scores for clean and noisy training samples on four datasets under 50\% IDN. A larger score indicates a higher estimated risk of label corruption.}
\label{fig:audit_distribution}
\end{figure*}

As shown in Fig.~\ref{fig:audit_distribution}, under 50\% IDN noisy samples receive clearly higher audit scores than clean samples, and the gap between their means ranges from 0.45 to 0.60 across the four datasets. In Fig.~\ref{fig:motivation}, we treat the corrupted labels as the positive class and compute the detection AUC after every epoch following warm-up, using GALA's audit score and one minus NLC's label-verification probability; the curves report the mean over five runs. GALA's detection AUC rises during training, whereas NLC's declines, indicating that the anchor-based evidence stays informative even as the model fits the noisy labels.

\subsection{Parameter Sensitivity}

\begin{figure*}[!t]
\centering
\includegraphics[width=\textwidth]{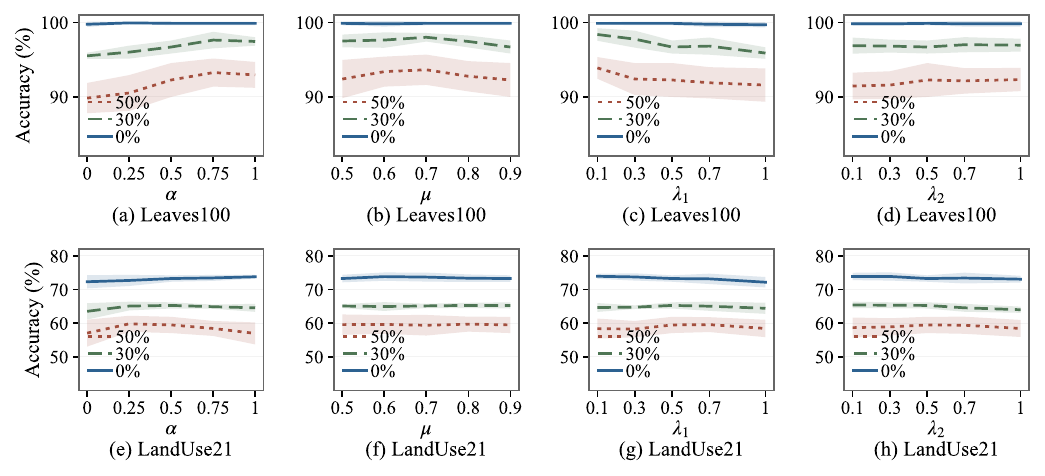}
\caption{Parameter sensitivity on Leaves100 (top) and LandUse21 (bottom) under different IDN rates. Shaded regions denote the standard deviation over five runs.}
\label{fig:param_sensitivity}
\end{figure*}

We study $\alpha$, $\mu$, $\lambda_1$, and $\lambda_2$ on Leaves100 and LandUse21 under 0\%, 30\%, and 50\% IDN. As shown in Fig.~\ref{fig:param_sensitivity}, $\alpha$ has a more visible effect under high noise, whereas its clean-data curves remain nearly flat. Performance changes little across different values of $\mu$ and $\lambda_2$, while $\lambda_1$ causes moderate variation under noisy labels. No parameter causes a sharp performance drop near the adopted values. We therefore use $\alpha=0.5$, $\mu=0.9$, and $\lambda_1=\lambda_2=0.5$ throughout the experiments.

\begin{figure*}[!t]
\centering
\includegraphics[width=\textwidth]{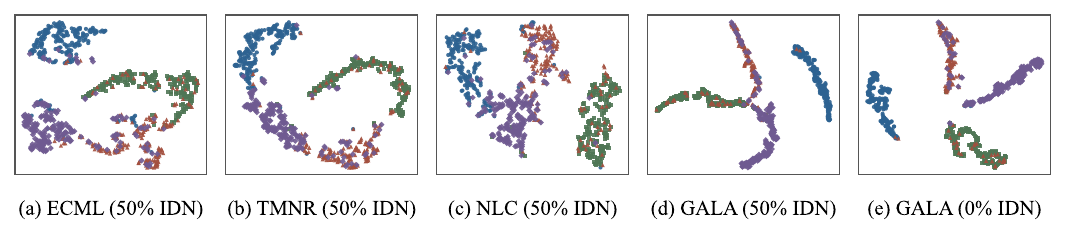}
\caption{t-SNE visualization on Digit4k using the same data split, random seed, and t-SNE configuration. The first four panels are trained under 50\% IDN, while the last panel shows GALA trained with clean labels.}
\label{fig:tsne}
\end{figure*}

\subsection{Ablation Study}

Under 50\% IDN, we compare the full GALA with four variants. GALA-1 replaces anchor auditing with NLC-style neighbor-based detection, which scores each label by how much the predictions of its 20 nearest neighbors disagree with the given label. GALA-2 removes label correction but retains sample re-weighting. GALA-3 freezes the anchors after initialization. GALA-4 removes both contrastive losses.

As shown in Table~\ref{tab:ablation}, the full model achieves the highest average accuracy of 79.29\% and performs best on five of the six datasets. Replacing anchor auditing with neighbor-based detection in GALA-1 lowers the average by 2.60 points, supporting the use of class-level anchors. GALA-2 causes the largest drop of 6.09 points, showing that re-weighting alone cannot recover the useful supervision in corrupted samples. Freezing the anchors in GALA-3 and removing both contrastive losses in GALA-4 reduce the average by 5.12 and 4.91 points. On BBC, the small gap between GALA and GALA-2 suggests that label correction helps less there than on the other datasets.

\begin{table}[!t]
\centering
\small
\setlength{\tabcolsep}{1.1mm}
\begin{tabular}{@{}lccccc@{}}
  \toprule
Dataset & GALA-1 & GALA-2 & GALA-3 & GALA-4 & GALA \\
\midrule
LandUse21 & 55.43 & 59.25 & 53.95 & 57.48 & \textbf{59.48} \\
Leaves100 & 83.50 & 83.19 & 92.19 & 92.00 & \textbf{92.25} \\
Scene15 & 75.20 & 74.94 & 72.22 & 74.52 & \textbf{75.23} \\
BBC & 69.64 & \textbf{70.95} & 70.80 & 67.30 & 70.51 \\
Mfeat & 91.10 & 70.10 & 82.85 & 77.45 & \textbf{92.10} \\
Digit4k & 85.25 & 80.78 & 72.98 & 77.50 & \textbf{86.17} \\
\midrule
Average & 76.69 & 73.20 & 74.17 & 74.38 & \textbf{79.29} \\
  \bottomrule
\end{tabular}
\caption{Ablation results under 50\% IDN. Each entry is the average classification accuracy (\%) over five runs, and the best result in each row is indicated by boldface.}
\label{tab:ablation}
\end{table}

\subsection{Representation Visualization}

Fig.~\ref{fig:tsne} compares representations on Digit4k using the same split, seed, and t-SNE configuration. Under 50\% IDN, ECML, TMNR, and NLC show partially mixed classes: samples with corrupted labels are pulled toward the wrong classes and blur the boundaries between neighboring groups. In contrast, GALA forms more compact groups with clearer separation, because the prototypical contrastive loss pulls features toward their class anchors and the sample weights limit the influence of suspicious samples. Although some dispersion remains, the representation of GALA is much closer to the clean-label reference, in line with the accuracy results in Table~\ref{tab:main_results}.

\FloatBarrier

\section{Conclusion}

We presented GALA, a global-anchor-based framework for multi-view classification with noisy labels. Instead of relying on the changing predictions of individual neighbors, GALA audits each observed label by comparing the instance with its assigned and competing anchors in every view and combining this evidence with the classifier confidence. The audit score is used for soft sample weighting and conservative label correction, and the corrected labels update the anchors and guide representation learning. Across six benchmarks and six noise settings, GALA ranks first in 33 of 36 settings and outperforms the strongest baseline by 7.34 points on average at 50\% noise. Overall, global anchors and cross-view consensus offer a dependable reference under label noise.

\FloatBarrier
\Needspace{8\baselineskip}
\section*{Acknowledgments}
This work was supported by the National Natural Science Foundation of China
(Grant No. 12501719), the Fundamental Research Funds for the Central
Universities (Grant No. QNTD202510), and the Beijing Training Program of
Innovation and Entrepreneurship for Undergraduates (Grant No. 202510022420).

\bibliographystyle{abbrvnat}
\bibliography{ref}

\end{document}